\documentclass[runningheads]{llncs}
\usepackage[T1]{fontenc}
\usepackage{amsmath}
\usepackage{graphicx}
\usepackage{amssymb}
\usepackage{pifont}
\usepackage{multirow}
\usepackage{color, colortbl}
\usepackage{subcaption}
\usepackage{marvosym}
\definecolor{Gray}{gray}{0.9}
\usepackage[group-separator={,}]{siunitx} 

\usepackage{subcaption}
\usepackage{color, colortbl}
\newcommand{\xmark}{\tiny-}
\begin{document}
%
\title{Towards Holistic Surgical Scene Understanding}
%
\author{Natalia Valderrama\inst{1}$^{\textrm{(\Letter)}}$, Paola Ruiz Puentes$^1$\thanks{Equal contribution.}, Isabela Hernández$^1$$^\star$, Nicolás Ayobi\inst{1}, Mathilde Verlyck\inst{1}, Jessica Santander\inst{2}, Juan Caicedo\inst{2}, Nicolás Fernández\inst{3,4} \and Pablo Arbeláez\inst{1}$^{\textrm{(\Letter)}}$}

\authorrunning{N. Valderrama et al.}
%
\institute{Center for Research and Formation in Artificial Intelligence, \\
Universidad de los Andes, Bogotá, Colombia \and
Fundación Santafé de Bogotá, Bogotá, Colombia \and 
Seattle Children's Hospital, Seattle, USA \and
University of Washington, Seattle, USA \\
\email{\{nf.valderrama, p.arbelaez\}@uniandes.edu.co}}
\maketitle              
\begin{abstract}
Most benchmarks for studying surgical interventions focus on a specific challenge instead of leveraging the intrinsic complementarity among different tasks. 
In this work, we present a new experimental framework towards holistic surgical scene understanding. First, we introduce the Phase, Step, Instrument, and Atomic Visual Action recognition (PSI-AVA) Dataset. 
PSI-AVA includes annotations for both long-term (Phase and Step recognition) and short-term reasoning (Instrument detection and novel Atomic Action recognition) in robot-assisted radical prostatectomy videos. 
Second, we present Transformers for Action, Phase, Instrument, and steps Recognition (TAPIR) as a strong baseline for surgical scene understanding. TAPIR leverages our dataset's multi-level annotations as it benefits from the learned representation on the instrument detection task to improve its classification capacity. 
Our experimental results in both PSI-AVA and other publicly available databases demonstrate the adequacy of our framework to spur future research on holistic surgical scene understanding. Code and data are available at \url{https://github.com/BCV-Uniandes/TAPIR}.

\keywords{Holistic Surgical Scene Understanding \and  Robot-Assisted Surgical Endoscopy \and Dataset \and Vision Transformers}
\end{abstract}
\section{Introduction}
Surgeries can be generally understood as compositions of interactions between instruments and anatomical landmarks, occurring inside sequences of deliberate activities. 
In this respect, surgical workflow analysis aims to understand the complexity of procedures through data analysis.
This field of research is essential for providing context-aware intraoperative assistance \cite{surgical-data-science}, streamlining surgeon training, and as a tool for procedure planning and retrospective analysis \cite{paper-fases}. 
Furthermore, holistic surgical scene understanding is essential to develop autonomous robotic assistants in the surgical domain. 

This work addresses the comprehension of surgical interventions through a systematical deconstruction of procedures into multi-level activities performed along robot-assisted radical prostatectomies. 
We decompose the procedure into short-term activities performed by the instruments and surgery-specific long-term processes, termed phases and steps.
We refer to the completion of a surgical objective as a step, whereas phases are primary surgical events, and are composed of multiple steps.
Further, each step contains certain instruments that perform a set of actions to achieve punctual goals.
Stressing on their intrinsic complementarity, the spatio-temporal relationship between multi-level activities and instruments is essential for a holistic understanding of the scene dynamics.
Most of the currently available datasets for surgical understanding do not provide enough information to study whole surgical scenes, as they focus on independent tasks of a specific surgery~\cite{m2cai16,saras,jigsaws,or-dataset}. 
These task-level restrictions lead to approaches with a constrained understanding of surgical workflow.
Only a fraction of surgical datasets enable multi-level learning~\cite{endonet-cholec80,cholecT40,heico,misaw}, despite the proven success of multi-level learning in complex human activity recognition~\cite{moma}. 

\begin{figure}[t]
\centering
\includegraphics[width=0.75\linewidth]{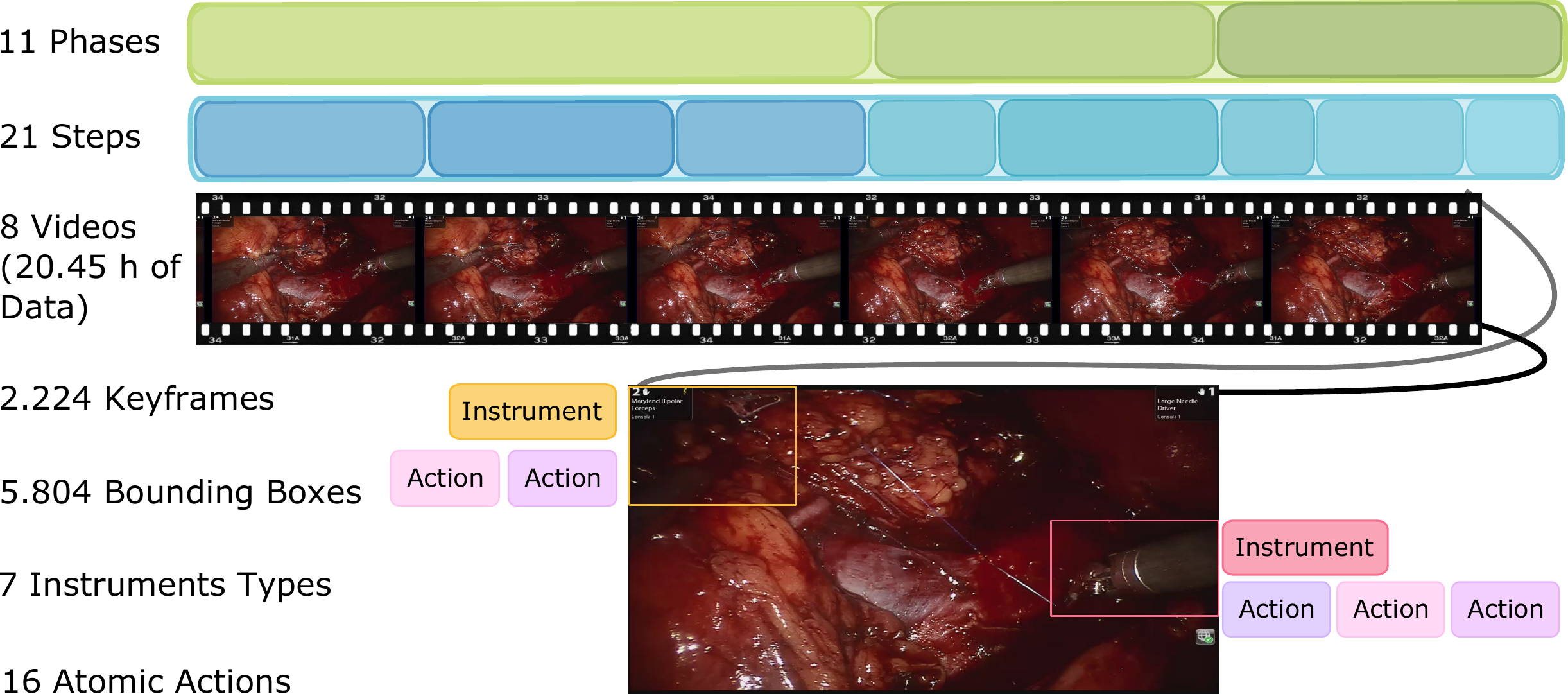}
\caption{\textbf{PSI-AVA Dataset} enables a holistic analysis of surgical videos through annotations for both long-term (Phase and Step Recognition) and short-term reasoning tasks (Instrument Detection and Atomic Action Recognition).}
\label{fig:dataset_overview}
\end{figure}

The first main contribution of this work is a novel benchmark for Phase, Step, Instrument, and Atomic Visual Action (PSI-AVA) recognition tasks for holistic surgical understanding (Figure~\ref{fig:dataset_overview}), which enables joint multi-task and multi-level learning for robot-assisted radical prostatectomy surgeries. 
Inspired by~\cite{AVA_2018}, we propose a surgical Atomic Action Recognition task where actions are not limited to a specific surgery nor dependent of interacted objects, and cannot be broken down into smaller activities. 

Table \ref{tab:rw_comparison} compares the main datasets for surgical workflow analysis.
Long-term surgical phase recognition is the most studied task and is frequently considered with instrument recognition. 
Finer action tasks fit in the context of gesture recognition through verb labels~\cite{saras,cholecT40,misaw,jigsaws,avos,nwoye2021rendezvous}, but are specific to one surgery type, thus motivating the added atomization of actions in our dataset. 
Current spatial annotations include instance-wise bounding boxes~\cite{saras,avos}, segmentation masks~\cite{heico,endovis2017,endovis2018} and kinematic data~\cite{jigsaws}. 
These annotations can be used to support coarser classification-based tasks. 
Lastly, multi-level learning is duly approached in~\cite{heico}, enabling the concurrent study of tasks in the outermost levels of surgical timescale analysis.
Our dataset extends this analysis to intermediate scales, solely using the more widespread video format while providing a solid experimental framework for each of the proposed tasks.

\begin{table}[t]
\centering
\caption{PSI-AVA compared against current surgical workflow analysis datasets. We indicate if the dataset has (\checkmark), does not have (--), has a different (*) version of the attribute, or information is not available (?).}
\label{tab:rw_comparison}
\resizebox{\textwidth}{!}{%
\begin{tabular}{l c c c c c c c c c}
\hline
Dataset & \begin{tabular}[c]{@{}c@{}}Multi-level\\ Understanding\end{tabular} & \begin{tabular}[c]{@{}c@{}}Phase\\ Recognition\end{tabular} & \begin{tabular}[c]{@{}c@{}}Step\\ Recognition\end{tabular} & \begin{tabular}[c]{@{}c@{}}Instrument\\ Recognition\end{tabular} & \begin{tabular}[c]{@{}c@{}}Instrument\\ Detection\end{tabular} & \begin{tabular}[c]{@{}c@{}}Action\\ Task\end{tabular} & \begin{tabular}[c]{@{}c@{}}Spatial\\ Annotations\end{tabular} & \begin{tabular}[c]{@{}c@{}}Video\\ hours\end{tabular} & \begin{tabular}[c]{@{}c@{}}Publicly\\ Available\end{tabular} \\ \hline \hline
LSRAO~\cite{or-dataset} & \xmark & \xmark & \xmark & \xmark & \xmark & \checkmark & \xmark & ? & \xmark \\
\rowcolor{Gray} JIGSAWS~\cite{jigsaws} & \xmark & \xmark & \xmark & \xmark & \xmark & \checkmark * & * & 2.62 & \checkmark \\
TUM LapChole~\cite{m2cai16} & \xmark & \checkmark & \xmark & \xmark & \xmark & \xmark & \xmark & 23.24 & \checkmark \\ \rowcolor{Gray}
Cholec80~\cite{endonet-cholec80} & \checkmark & \checkmark & \xmark & \checkmark & \xmark & \xmark & \xmark & 51.25 & \checkmark \\ 
\begin{tabular}[c]{@{}l@{}}CaDIS~\cite{cadis} \end{tabular} & \xmark & \checkmark & \xmark & \checkmark & \xmark & \xmark & \xmark & 9.1 & \checkmark \\ \rowcolor{Gray}
HeiCo~\cite{heico} & \checkmark & \checkmark & \xmark & \xmark & \checkmark & \xmark & \checkmark & 96.12 & \checkmark \\ 
CholecT40,CholecT50~\cite{cholecT40,nwoye2021rendezvous} & \xmark & \xmark & \xmark & \checkmark & \xmark & \checkmark & \xmark & ? & \xmark \\ \rowcolor{Gray}
MISAW~\cite{misaw} & \checkmark & \checkmark & \checkmark & \xmark {\normalsize *} & \xmark & \checkmark & \xmark & 1.52 & \checkmark \\
ESAD~\cite{saras} & \xmark & \xmark & \xmark & \xmark & \xmark & \checkmark & \checkmark & 9.33 & \checkmark \\ \rowcolor{Gray}
AVOS~\cite{avos} & \xmark & \xmark & \xmark & \checkmark & \checkmark & \checkmark & \checkmark & 47 & \xmark \\ \hline\hline
\textbf{PSI-AVA (Ours)} & \checkmark & \checkmark & \checkmark & \checkmark & \checkmark & \checkmark & \checkmark & 20.45 & \checkmark \\ 
\hline
\end{tabular}%
}
\end{table}

Recently, Vision Transformers have achieved outstanding performance in both natural image and video related tasks~\cite{detr,def_detr,vit,mvit}, stressing on the benefit of analyzing complex scenes through attention mechanisms.  
Specifically, transformers can exploit general temporal information in sequential data, providing convenient architectures for surgical video analysis, and overcoming the narrow receptive fields of traditional CNNs.
Recent methods leverage transformers' attention modules and achieve outstanding results in phase recognition~\cite{opera,transvnet,ding2021exploiting}, interaction recognition~\cite{nwoye2021rendezvous}, action segmentation~\cite{towards} and tools detection, and segmentation~\cite{trasetr,lapformer}. 
Additionally,~\cite{nwoye2021rendezvous} proposes a weakly supervised method for instrument-action-organ triplets recognition. 
Nonetheless, no previous research has employed transformers for surgical atomic action recognition. 

\begin{figure}[t]
\centering
\includegraphics[width=0.9\linewidth]{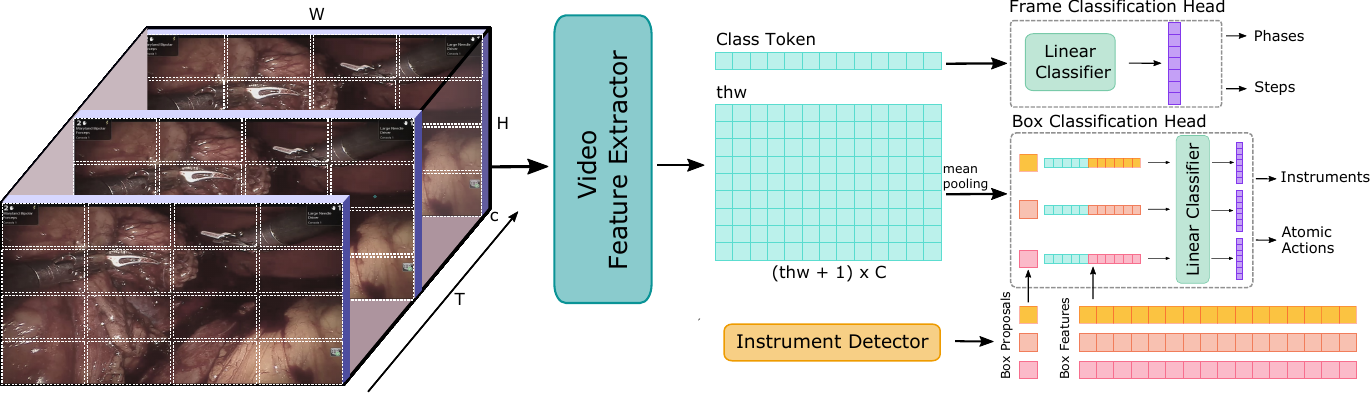}
\caption{\textbf{TAPIR}. Our approach leverages the global, temporal information extracted from a surgical video sequence with localized appearance cues. The association of complementary information sources enables the multi-level reasoning over dynamic surgical scenes. Best viewed in color.}
\label{fig:method_overview}
\end{figure}

As a second main contribution towards holistic surgical scene understanding, we introduce the Transformers for Action, Phase, Instrument, and steps Recognition (TAPIR) method. 
Figure~\ref{fig:method_overview} shows our general framework that consists of two main stages: a video feature extractor followed by a task-specific classification head.
Our method benefits from box-specific features extracted by an instrument detector to perform the more fine-grained recognition task. 
TAPIR performance on the PSI-AVA dataset highlights the benefits of multi-level annotations and its relevance to holistic understanding on surgical scenes. 

Our main contributions can be summarized as follows: (1) We introduce and validate the PSI-AVA dataset with annotations for phase and step recognition, instrument detection, and the novel task of atomic action recognition in surgical scenes. (2) We propose TAPIR, a transformer-based method that leverages the multi-level annotations of PSI-AVA dataset and establishes a baseline for future work in our holistic benchmark.

To ensure our results' reproducibility and promote further research on holistic surgical understanding, we will make publicly available the PSI-AVA dataset, annotations, and evaluation code, as well as pre-trained models and source code for TAPIR under the MIT license at \url{https://github.com/BCV-Uniandes/TAPIR}.

\section{Data and Ground-Truth Annotations}
The endoscopic videos were collected at Fundación Santafé de Bogotá during eight radical prostatectomy surgeries, performed with the Da Vinci SI3000 Surgical System. 
Data collection and post-processing protocols were done under approval of the ethics committees of all institutions directly involved in this work. 
Regarding the experimental framework, we formulate a two-fold cross-validation setup. Each fold contains data related to four procedures and a similar distribution of phase, step, instrument, and atomic action categories. Each surgery is uniquely assigned to one of the folds. Our dataset contains annotations for four tasks that enable a comprehensive evaluation of surgical workflow at multiple scales.
Next, we describe each of the tasks, their annotation strategies, and evaluation methodology, followed by a report of dataset statistics. 

\textbf{Phase and Step Recognition}. 
These tasks share the common objective of identifying surgical events, but differ in the temporal scale at which these events occur. 
Annotations for these frame-level tasks were done by two urologist experts in transperitoneal radical prostatectomy surgeries. 
The defined activity partonomy includes 11 phases and 20 steps, both including an \textit{Idle} category to indicate periods of instrument inactivity.
See Supplementary Figure 1 for more information regarding the phase and step categories.
To assess performance in these frame-level tasks, we use the mean Average Precision (mAP), a standard metric for action recognition in videos \cite{AVA_2018}.

\textbf{Instrument Detection}. 
This task refers to the prediction of instrument-level bounding boxes and corresponding category labels, from a set of 7 possible classes. 
A hybrid strategy was carried out to annotate all bounding boxes in our dataset's keyframes.
A domain-specific Faster R-CNN model pretrained on the task of instrument detection on the EndoVis 2017~\cite{endovis2017} and 2018~\cite{endovis2018} datasets is used to generate initial bounding box estimates. 
Next, crowd-source annotators verify and refine the bounding box predictions.
Annotators are instructed to verify that all bounding boxes enclose complete instruments, and to select one category label per bounding box.
The evaluation of this task uses the mAP metric at an Intersection-over-Union threshold of 0.5 (mAP@0.5 IoU).

\textbf{Atomic Action Recognition}. 
Once the spatial location of instruments is defined, the aim of this novel task is to describe their localized activity by fine action categories. 
Similarly to~\cite{AVA_2018}, our atomic action vocabulary is based on two principles: generality and atomicity. 
The annotations for this task associate the previously defined bounding boxes to a maximum of three atomic actions per instrument, from a set of 16 classes. 
To evaluate this task, we employ the standard metric for action detection in video: mAP@0.5 IoU.

\textbf{Dataset Statistics.} After data collection, an anonymization and curation process was carried out, removing video segments with non-surgical content and potential patient identifiers. 
As a result, our dataset contains approximately 20.45 hours of the surgical procedure performed by three expert surgeons.
Given that phase and step annotations are defined over fixed time intervals, their recognition tasks benefit from dense, frame-wise annotations at a rate of one keyframe per second. 
As a result, \num{73618} keyframes are annotated with one phase-step pair.
Due to the fine-grained nature of instruments and atomic actions annotations, we sample frames every 35 seconds of video, to obtain \num{2238} keyframe candidates.
These keyframes are annotated with \num{5804} instrument instances, with corresponding instrument category and atomic action(s). 
Supplementary Figure 2 depicts the distribution of instances within the different categories per task.
We highlight the imbalanced class distribution in the proposed tasks inherent in surgical-based problems. 
The less frequent classes in our dataset correspond to sporadic and case-specific surgical events.


\section{Method}  \label{sec:method} 

\subsection{TAPIR}
We propose TAPIR: a Transformer model for Action, Phase, Instrument, and step Recognition in robot-assisted radical prostatectomy surgery videos. Figure~\ref{fig:method_overview} depicts an overview of this baseline method.
We build upon the Multiscale Vision Transformer (MViT) model developed for video and image recognition tasks~\cite{mvit}.
The connection between multiscale feature hierarchies with a transformer model, provides rich features to model the complex temporal cues of the surgical scene around each keyframe in PSI-AVA.
The extraction of these features constitutes the first stage of TAPIR, followed by a frame classification head for phase and step recognition.
Supplementary Figure 3 provides a detailed architecture of the Video Feature Extractor stage of our method.
To account for instrument detection and atomic action recognition tasks, we employ the Deformable DETR \cite{def_detr} method to obtain bounding box estimates and box-specific features, which are combined with the keyframe features by an additional classification head to predict instrument category or atomic action(s).
The following subsections provide further details on the methods we build upon.

\textbf{Instrument Detector.} Deformable DETR \cite{def_detr} is a transformer-based detector that implements multi-scale deformable attention modules to reduce the complexity issues of transformer architectures. 
This model builds upon DETR \cite{detr}, and comprises a transformer encoder-decoder module that predicts a set of object detections based on image-level features extracted by a CNN backbone. Furthermore, this method applies an effective iterative bounding
box refinement mechanism, to enhance the precision of the bounding boxes coordinates.

\textbf{Video Feature Extractor.} MViT \cite{mvit} is a transformer-based architecture for video and image recognition that leverages multiscale feature hierarchies. 
This method is based on stages with channel resolution rescaling processes. 
The spatial resolution decreases from one stage to another while the channel dimension expands. 
A Multi-Head Pooling Attention (MHPA) pools query, key, and value tensors to reduce the spatiotemporal resolution.
This rescaling strategy produces a multiscale pyramid of features inside the network, effectively connecting the principles of transformers with multi-scale feature hierarchies, and allows capturing low-level features at the first stages and high-level visual information at deeper stages. 

\textbf{Classification Heads.} Depending on the granularity of the task, we use a frame classification head or a box classification head. 
The former receives the features from the Video Feature Extractor.
As described in \cite{mvit}, MViT takes the embedding of the token class and performs a linear classification into the phase or step category. 
The box classification head receives the features extracted from the video sequence and the box-specific features from the Instrument Detector. 
After an average pooling of the spatiotemporal video features, we concatenate this high-level frame information with the specific box features. 
Lastly, a linear layer maps this multi-level information to an instrument or atomic action class.

\textbf{Implementation Details.} We pretrained Deformable DETR \cite{def_detr} with a ResNet-50 \cite{resnet} backbone on the EndoVis 2017 \cite{endovis2017} and EndoVis 2018 dataset \cite{endovis2018} (as defined in \cite{isinet}). 
We train our instrument detection with the iterative bounding box refinement module in PSI-AVA for 100 epochs in a single Quadro RTX 8000 GPU, and a batch size of 6. 
As in \cite{slowfast,mvit}, we select the bounding boxes with a confidence threshold of 0.75.

\begin{table}[b]
\centering
\caption{Performance on PSI-AVA dataset. We report the number of parameters, inference cost in FLOPs, and mean Average Precision (mAP) on our two-fold cross-validation setup along with the standard deviation.}\label{tab:main_results}
\resizebox{\textwidth}{!}{%
\begin{tabular}{llcc|ccl|cc}
\hline
\multicolumn{2}{c}{\multirow{2}{*}{Method}} & \multirow{2}{*}{FLOPs (G)} & \multirow{2}{*}{Param (M)} & \multicolumn{3}{c|}{Long-term Tasks mAP (\%)} & \multicolumn{2}{c}{Short-term Tasks mAP@0.5IoU (\%)} \\ \cline{5-9} 
\multicolumn{2}{c}{} & & & Phases & \multicolumn{2}{c|}{Steps} & Instruments & Atomic Actions \\ \hline
\multicolumn{2}{l}{SlowFast \cite{slowfast}} & 81.24 & 33.7 & \multicolumn{1}{c}{46.80 ± 2.07} & \multicolumn{2}{c|}{33.86 ± 1.93} & \multicolumn{1}{c}{80.29 ± 0.92} & \multicolumn{1}{c}{18.57 ± 0.62} \\
\multicolumn{2}{l}{TAPIR-VST} & 65.76 & 87.7 & \multicolumn{1}{c}{51.63 ± 0.89} & \multicolumn{2}{c|}{41.00 ± 3.34} & \multicolumn{1}{c}{\textbf{81.14 ± 1.51}} & \multicolumn{1}{c}{22.49 ± 0.34} \\
\multicolumn{2}{l}{TAPIR (Ours)} & 70.80 & 36.3 & \multicolumn{1}{c}{\textbf{56.55 ± 2.31}} & \multicolumn{2}{c|}{\textbf{45.56 ± 0.004}} & \multicolumn{1}{c}{80.85 ± 1.54}  & \multicolumn{1}{c}{\textbf{28.68 ± 1.33}}    \\
\hline
\end{tabular}}
\end{table}
As Video Feature Extractor, we adapt and optimize MViT-B pretrained on Kinetics-400 \cite{kinetics} for the long-term recognition tasks. For the instrument detection and atomic action recognition tasks, we optimized an MViT model pretrained for the step recognition task.
For video analysis, we select a 16-frame sequence centered on an annotated keyframe with a sampling rate of 1. We train until convergence, with a 5 epoch warm-up period and a batch size of 6 for long-term tasks and 24 for short-term tasks. We employ a binary cross-entropy loss for the atomic action recognition task and a cross-entropy loss for the remaining tasks, all with an SGD optimizer, a cosine scheduler with a weight decay of $1e^{-4}$, and a base learning rate of $0.125$ for the atomic actions task, and $0.0125$ for the rest of the tasks.

\section{Experiments}
\subsection{Video Feature Extractor method comparison}

\textbf{Deep Convolutional Neural Networks-based (DCNN) method.}
Similar to the methodology described in Section \ref{sec:method}, we replace the Instrument Detector and the Video Feature Extractor based on transformers with two DCNN methods. 
We use Faster R-CNN as detector \cite{ren2015faster}, which we pretrain on the EndoVis 2017 dataset \cite{endovis2017} and the training set of the 2018 dataset \cite{endovis2018} (as described in \cite{isinet}). 
We train on PSI-AVA for 40 epochs, with a batch size of 16 in 1 GPU, a base learning rate of 0.01, and use a SGD optimizer. 
Finally, we keep the predictions with a confidence score higher than \num{0.75}. Additionally, as video feature extractor we use SlowFast \cite{slowfast} pretrained on CaDIS \cite{cadis} and likewise to TAPIR we select clips of 16 consecutive frames centered on an annotated keyframe.
We set the essential SlowFast parameters $\alpha$ and $\beta$ to 8 and $\frac{1}{8}$, respectively. 
We train for 30 epochs, with a 5 epoch warm-up period and a batch size of 45. We employ SGD optimizer, a base learning rate of $0.0125$, and a cosine scheduler with a weight decay of $1e^{-4}$. 
For instrument detection and atomic action recognition tasks we perform a learnable weighted sum of the instrument detector features and SlowFast temporal features, and we apply a linear classification layer to obtain their corresponding predictions.

\textbf{TAPIR-VST.}
To complement our study, we compare the use of MViT as Video Feature Extractor with another state-of-the-art transformer method: Video Swin Transformer \cite{vst} (VST). We follow the methodology and training curriculum described in Section 3, replacing MViT in TAPIR for Video Swin Transformer with Swin-B as a backbone, pretrained on ImageNet-22K. We keep Deformable DETR as the Instrument Detector.

\subsection{Results and Discussion}
Table \ref{tab:main_results} shows a comparison study between transformer-based and DCNN-based methods on the PSI-AVA dataset. TAPIR and TAPIR-VST achieve a significant improvement in all tasks over the DCNN-based method. 
However, TAPIR overcomes TAPIR-VST in the three recognition tasks with a lower computational cost.
Our best method outperforms SlowFast \cite{slowfast} by 11.7 \%, and 9.75 \% in mAP for steps and phases recognition tasks, respectively. 
Furthermore, TAPIR improves 10.11\% mAP@0.5IoU in our challenging atomic action recognition task, stressing on the relevance of understanding the whole scene to accurately identify the actions performed by each instrument.
Qualitative comparisons of SlowFast and TAPIR in all tasks are shown in Supplementary Figure 4.
Overall, these results demonstrate the superior capacity of transformer-based methods to understand the surgical workflow in video and a higher recognition capacity of atomic actions. Also, indicate that TAPIR represents a strong baseline for future research on surgical workflow understanding.

\textbf{PSI-AVA validation.} 
To validate PSI-AVA's benchmark, we study the behavior of the proposed baselines in our dataset and other public databases whose task definition is equivalent to ours. 
We compare our image-level recognition tasks with their counterparts in the MISAW \cite{misaw} dataset, and compare our instrument detection task with EndoVis 2017 \cite{endovis2017} dataset.
\begin{table}[t]
\caption{Comparison of PSI-AVA with publicly available datasets. We report the number of parameters, inference cost in FLOPs, and mAP performance with corresponding standard deviations.}
\label{tab:dataset_comp}
\resizebox{1\textwidth}{!}{
\begin{tabular}{lcc|cccc}
\hline
\multicolumn{1}{c}{\multirow{2}{*}{Method}} & \multirow{2}{*}{FLOPs (G)} & \multirow{2}{*}{Param. (M)} & \multicolumn{2}{c|}{Phase Recognition Task mAP (\%)} & \multicolumn{2}{c}{Step Recognition Task mAP (\%)} \\ \cline{4-7} 
\multicolumn{1}{c}{} &  &  & MISAW \cite{misaw} & \multicolumn{1}{c|}{PSI-AVA} & MISAW \cite{misaw} & PSI-AVA \\ \hline
SlowFast \cite{slowfast} & 81.24 & 33.7 & 94.18 & \multicolumn{1}{c|}{46.80 ± 2.07} & 76.51 & 33.86 ± 1.93 \\
TAPIR (Ours) & 70.80 & 36.3 & \textbf{94.24} & \multicolumn{1}{c|}{\textbf{56.55 ± 2.31}} & \textbf{79.18} & \textbf{45.56 ± 0.004} \\ \hline \hline
\multicolumn{1}{c}{\multirow{2}{*}{Method}} & \multirow{2}{*}{FLOPs (G)} & \multirow{2}{*}{Param. (M)} & \multicolumn{4}{c}{Instruments Detection Task mAP@0.5IoU (\%)} \\ \cline{4-7} 
\multicolumn{2}{c}{} & &  \multicolumn{2}{c}{Endovis 2017 \cite{endovis2017}} & \multicolumn{2}{c}{PSI-AVA} \\ \hline
Faster R-CNN \cite{ren2015faster} & 180 & 42 & \multicolumn{2}{c}{47.51 ± 6.35} & \multicolumn{2}{c}{80.12 ± 2.32} \\
Def. DETR \cite{def_detr} & 173 & 40.6 & \multicolumn{2}{c}{\textbf{61.71 ± 3.79}} & \multicolumn{2}{c}{\textbf{81.19 ± 0.87}} \\
\hline
\end{tabular}} 
\end{table}
Table \ref{tab:dataset_comp} summarizes the results of the benchmark validation. 
The consistency in the relative order of the two families of methods across datasets provides empirical support for the adequacy of PSI-AVA as a unified testbed for holistic surgical understanding. 
Furthermore, the superiority of TAPIR and its instrument detector across datasets further support our method as a powerful baseline for multi-level surgical workflow analysis.

\textbf{Multi-Level Annotations allow better understanding.} In this set of experiments, we study the complementarity between the task of detection and classification, specifically instrument detection and atomic action recognition. 
Such complementarity is inherent to the concept of surgical scene understanding and supports the benefits of having multi-task annotations in PSI-AVA.
\begin{table}[t]
\centering
\caption{TAPIR results for detection-based tasks on PSI-AVA dataset. We report mAP performance on our cross-validation setup with standard deviation.}
\label{tab:class_detect}
\resizebox{0.8\textwidth}{!}{%
\centering
\begin{tabular}{cc|cc}
\hline
\multirow{2}{*}{\begin{tabular}[c]{@{}c@{}}Video Feature\\  Extractor\end{tabular}} & \multirow{2}{*}{\begin{tabular}[c]{@{}c@{}}Box Features \\ in Classification Head\end{tabular}} & \multicolumn{2}{c}{mAP@0.5IoU (\%)} \\ \cline{3-4} 
 &  & Instrument Detection & Action Recognition \\ \hline
 & W/o features & 13.96 ± 0.33 & 9.69 ± 0.07 \\
TAPIR (Ours) & Faster R-CNN & 80.40 ± 0.63 & 21.18 ± 0.35 \\
 & Deformable DETR & \textbf{80.85 ± 1.54} & \textbf{28.68 ± 1.33} \\
\hline
\end{tabular}
}
\end{table}
Table \ref{tab:class_detect} shows that TAPIR effectively exploits the complementary, multi-task, and multi-level information.
Without instrument-specific features, TAPIR's head classifies from the high-level temporal representation and loses the spatially localized information.
Adding instrument-specific features improves significantly the performance of both instrument detection and atomic action recognition tasks, which entails that the box classification head is capable of leveraging localized and temporal information. 
Consequently, higher performance in instrument detection, related to more instrument-specific features, reflects in a more accurate recognition by TAPIR (Table \ref{tab:class_detect}).

\section{Conclusions} 
In this work, we present a novel benchmark for holistic, multi-level surgical understanding based on a systematical deconstruction of robot-assisted radical prostatectomies.
We introduce and validate PSI-AVA, a novel dataset that will allow the study of intrinsic relationships between phases, steps and instrument dynamics, for which we introduce the atomic action definition and vocabulary. 
Furthermore, we propose TAPIR as a strong transformer-based baseline method tailored for multi-task learning.
We hope our framework will serve as a motivation for future work to steer towards holistic surgery scene understanding.

\section{Acknowledgements}

The authors would like to thank Laura Bravo-Sánchez, Cristina González and Gabriela Monroy for their contributions along the development of this research work. Natalia Valderrama, Isabela Hernández and Nicolás Ayobi acknowledge the support of the 2021 and 2022 UniAndes-DeepMind Scholarships. 

%
%
%
\bibliographystyle{splncs04}
\bibliography{bibliography}
\end{document}


%
\title{Towards Holistic Surgical Scene Understanding}
\author{Anonymous}
\institute{Anonymous Organization \\
\email{**@******.***}}

\subsubsection{}\textbf{Supplemental Material} Towards Holistic Surgical Scene Understanding.
\begin{figure}[h!]
\centering
\includegraphics[width=0.73\linewidth]{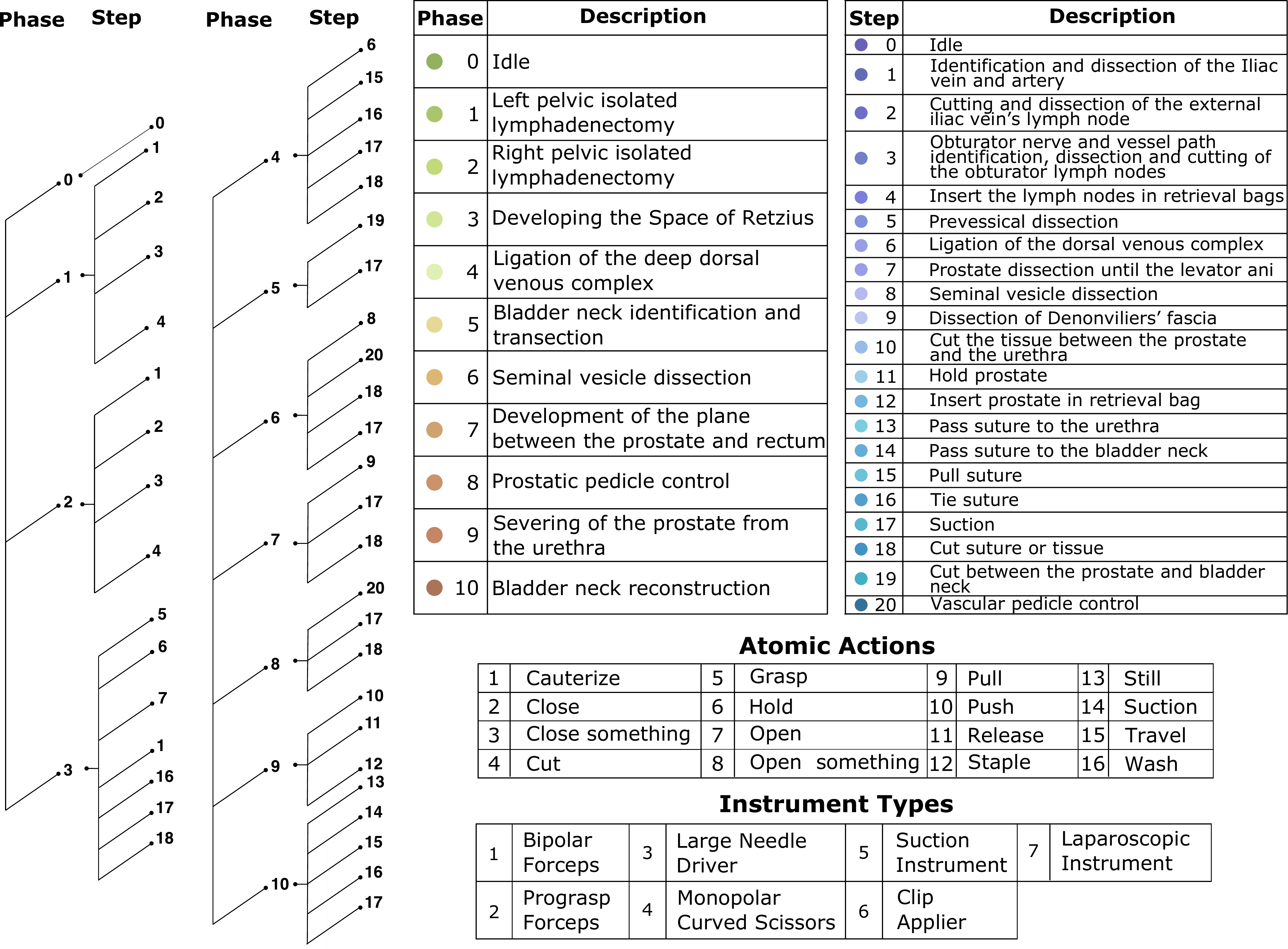}
\caption{\textbf{PSI-AVA classes per task}. (Left) the phases and steps are organized following the order found in a prostatectomy procedure. (Right) list of the class labels for the phase, step and atomic action recognition tasks and the instrument detection task. Best viewed in color.}
\label{sup-fig:partonomy}
\end{figure}

\begin{figure}[h!]
\centering
\includegraphics[width=0.73\linewidth]{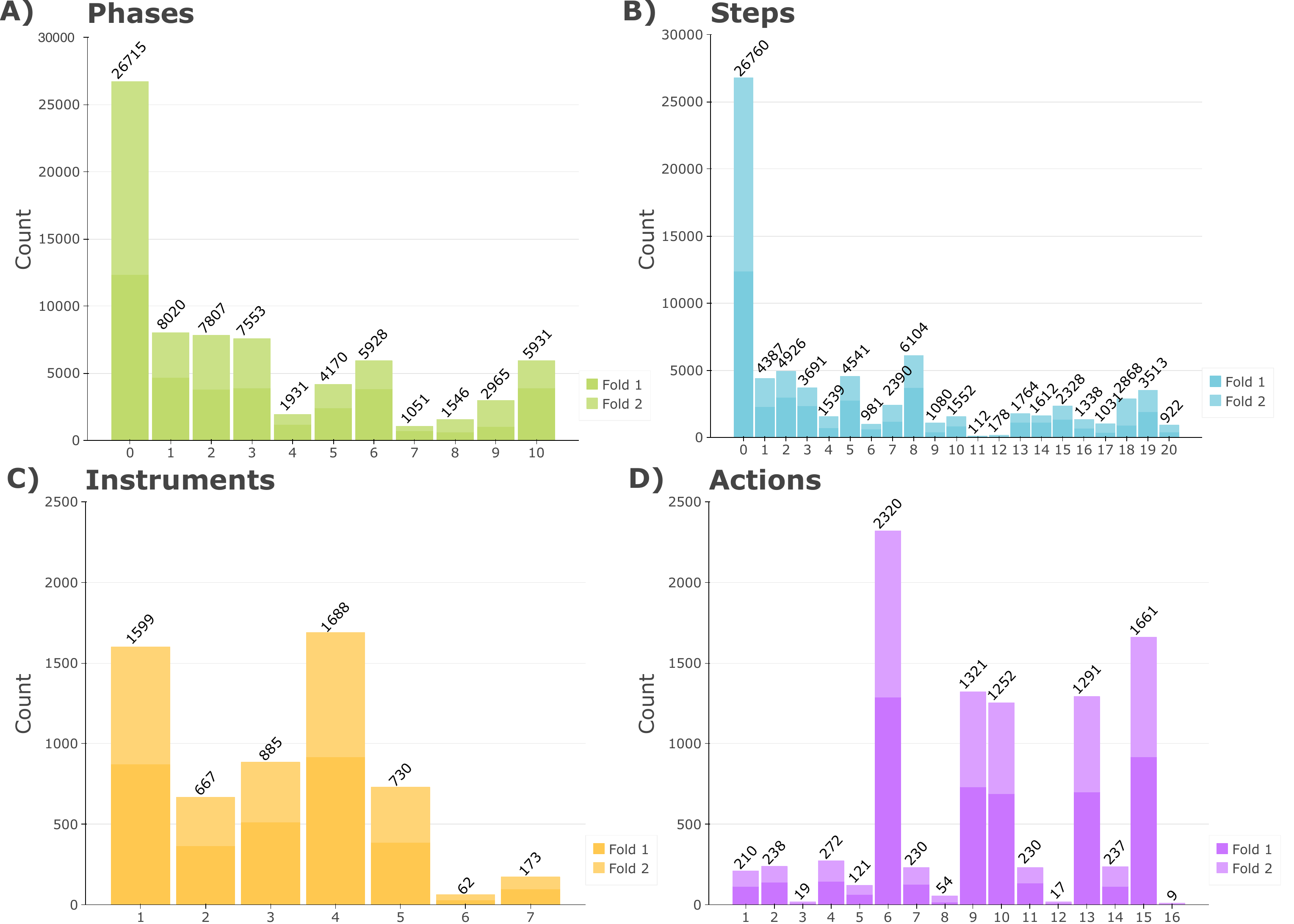}
\caption{\textbf{PSI-AVA Statistics}. Number of annotations for each class of the recognition and detection tasks. Colors denote the distribution in the fold partition.}
\label{sup-fig:statistics}
\end{figure}

\begin{figure}[h!]
\centering
\includegraphics[width=0.72\linewidth]{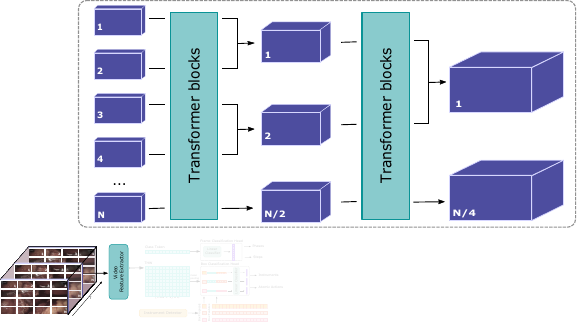}
\caption{\textbf{Video Feature Extractor architecture.} TAPIR builds upon MViT [9], which uses a multiscale pyramidal strategy to extract low-spatial but high-dimensional features from video sequences.}
\label{sup-fig:statistics}
\end{figure}

\begin{figure}[h!]
\centering
\includegraphics[width=0.7\linewidth]{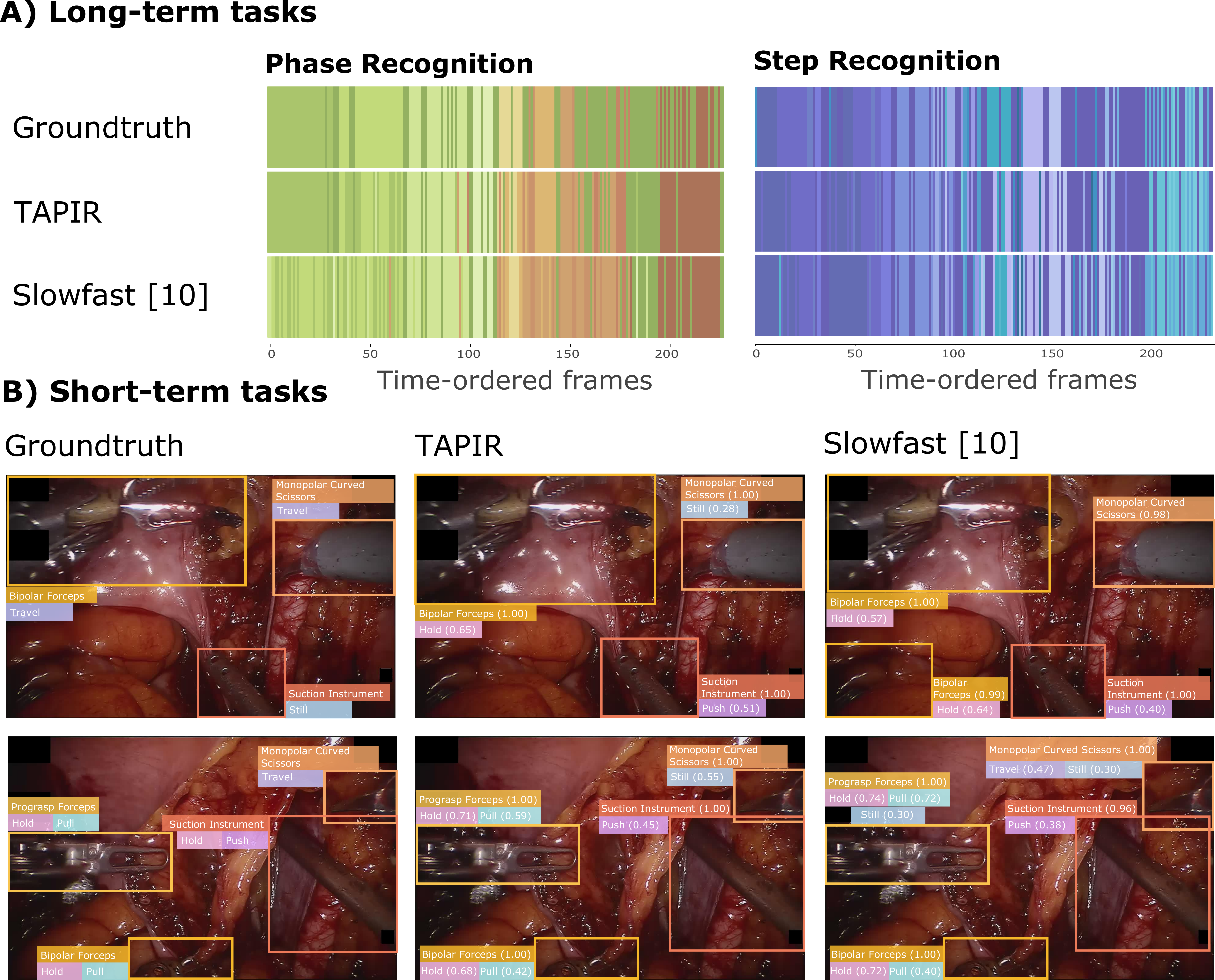}
\caption{\textbf{Performance comparison} between TAPIR and SlowFast [10] grouped by A) long-term and B) short-term tasks. A) For the Phase and Step Recognition tasks, TAPIR shows higher continuity along with its predictions, while SlowFast fails to keep coherence. Supplemental Figure 1 shows color codes for both tasks. B) Both methods fail to recognize some of the atomic actions, demonstrating the task's difficulty. However, TAPIR action prediction keeps coherence between the options, contrary to SlowFast's (e.g., travel and still). Best viewed in color.}
\label{sup-fig:statistics}
\end{figure}